\documentclass{article}

\usepackage{arxiv}

\usepackage[utf8]{inputenc} 
\usepackage[T1]{fontenc}    
\usepackage{natbib}         
\usepackage{hyperref}       
\usepackage{url}            
\usepackage{amssymb}        
\usepackage{booktabs}       
\usepackage{amsfonts}       
\usepackage{nicefrac}       
\usepackage{microtype}      
\usepackage{xcolor}         
\usepackage{upquote}
\usepackage{fancyvrb}
\usepackage{float}
\usepackage{xspace}      

\usepackage{soul}
\definecolor{deepblue}{rgb}{0,0,0.5}
\definecolor{deepred}{rgb}{0.6,0,0}
\definecolor{deepgreen}{rgb}{0,0.5,0}

\usepackage{listings}
\usepackage{graphicx}
\usepackage{courier}

\definecolor{mygreen}{rgb}{0,0.6,0}
\definecolor{mygray}{rgb}{0.5,0.5,0.5}
\definecolor{mymauve}{rgb}{0.58,0,0.82}
\DeclareFixedFont{\ttb}{T1}{txtt}{bx}{n}{10} 
\DeclareFixedFont{\ttm}{T1}{txtt}{m}{n}{10}  

\DeclareFixedFont{\ttbsmall}{T1}{txtt}{bx}{n}{8} 
\DeclareFixedFont{\ttmsmall}{T1}{txtt}{m}{n}{8}  

\newcommand\pythonstyle{\lstset{
language=Python,
basicstyle=\ttm,
morekeywords={self},              
keywordstyle=\ttb\color{deepblue},
emph={MyClass,__init__},          
emphstyle=\ttb\color{deepred},    
stringstyle=\color{deepgreen},
frame=tb,                         
showstringspaces=false
}}

\lstnewenvironment{python}[1][]
{
\pythonstyle
\lstset{#1}
}
{}

\newcommand\pythoninline[1]{{\pythonstyle\lstinline!#1!}}

\newcommand\pythonsmallstyle{\lstset{
language=Python,
basicstyle=\ttmsmall,
morekeywords={self},              
keywordstyle=\ttbsmall\color{deepblue},
emph={MyClass,__init__},          
emphstyle=\ttbsmall\color{deepred},    
stringstyle=\color{deepgreen},
frame=tb,                         
showstringspaces=false
}}

\lstnewenvironment{pythonsmall}[1][]
{
\pythonsmallstyle
\lstset{#1}
}
{}

\newcommand\bashstyle{\lstset{
language=bash,
basicstyle=\ttm,
morekeywords={},              
keywordstyle=\ttm,
emph={},          
emphstyle=\ttb\color{deepred},    
stringstyle=\color{deepgreen},
frame=tb,                         
showstringspaces=false
}}

\lstnewenvironment{bash}[1][]
{
\bashstyle
\lstset{#1}
}
{}

\newcommand\bashinline[1]{{\bashstyle\lstinline!#1!}}

\newcommand{\numdatasets}{338\xspace}
\newcommand{\numtrajectories}{350K\xspace}
\newcommand{\numsteps}{55M\xspace}
\newcommand{\datasetsize}{56\,GB\xspace}

\title{Minari: Infrastructure for Collaborative Reinforcement Learning Datasets}

\renewcommand{\shorttitle}{Minari}

\author{%
  Omar G.~Younis \\
  Farama Foundation, Mila \\
  \texttt{omar@farama.org} \\
  \And
  Rodrigo Perez-Vicente \\
  Farama Foundation \\
  \texttt{rperezvicente@farama.org} \\
  \And
  John U.~Balis \\
  Farama Foundation, Mila \\
  \texttt{balisujohn@gmail.com} \\
  \AND
  Alex Davey \\
  Farama Foundation, INRIA \\
  \texttt{alex.davey@inria.fr} \\
  \And
  Will Dudley \\
  Farama Foundation \\
  \texttt{william.dudley19@imperial.ac.uk} \\
  \And
  Jordan K.~Terry \\
  Farama Foundation \\
  \texttt{jkterry@farama.org} \\
}

\begin{document}

\maketitle

\begin{abstract}
Offline reinforcement learning (RL) depends on shared datasets, but sharing them is still hard. Each benchmark ships its own format, documentation is thin, and most datasets cannot be regenerated once the environment code moves on. We introduce \emph{Minari}, an open-source Python library for creating, sharing, and loading RL datasets. Its API follows Gymnasium, so recording a dataset means wrapping an environment and stepping through it as usual. From there, users can attach metadata, filter, combine, and shuffle datasets, and publish or download them with one command through the HuggingFace Hub. Every dataset stores the environment specification, seeds, applied wrappers, and pinned package requirements needed to rebuild the environment and regenerate the data.
TorchRL, AgileRL, and d3rlpy already load Minari datasets natively, so switching datasets in those libraries takes one line of code. Minari also hosts \numdatasets reproducible datasets (\numtrajectories trajectories, \numsteps steps) covering robotic locomotion and manipulation, maze and grid navigation, language-conditioned instruction following, Atari, and web navigation, including a full regeneration of D4RL with public generation scripts. We report baseline scores for six offline RL algorithms on the regenerated D4RL and MuJoCo datasets.
\end{abstract}


\section{Introduction}

Agents that learn from their own interaction data are an increasingly central part of AI research \citep{experience-era}. That data, the record of what an agent saw, did, and got in return, is expensive to collect and useful well beyond the project that produced it. Yet very little of it is shared in a form that others can reuse.

Three problems get in the way. First, there is no common format. Each benchmark defines its own layout for observations, actions, and episode boundaries, so a loader written for one rarely works on another. Second, released datasets are usually frozen snapshots: the environment version, seeds, and wrappers used to generate them are not recorded, and once the simulator changes the data can no longer be regenerated or evaluated under matching conditions. D4RL \citep{d4rl}, the most widely used offline RL benchmark, is a case in point; its datasets depend on deprecated versions of MuJoCo and Gym. Third, combining data from several sources means reconciling spaces, metadata, and conventions by hand.

Minari is our answer to these problems. It is an open-source library that defines a dataset format for sequential decision-making and the tools to create, share, and load data in that format. The API mirrors Gym and Gymnasium \citep{gym, gymnasium}, which most RL practitioners already use daily: Figure~\ref{fig:code} shows that recording a dataset takes a wrapper and one function call, and loading one takes a single line. Datasets are versioned and hosted on the HuggingFace Hub\footnote{\url{https://huggingface.co}}, and can be listed, downloaded, inspected, and combined from the command line or from Python. Section~\ref{sec:library_features} describes these features in detail.

We want Minari to do for RL datasets what Gymnasium did for environments: give the community one interface, so that a new dataset is usable by every compatible training library the day it is published.

\paragraph{Contributions.} This paper makes the following contributions:

\begin{itemize}
    \item \textbf{A dataset format and library.} An environment-agnostic format for trajectory data with machine-readable metadata and a fixed naming scheme, implemented in a pip-installable library with a Python API and a command-line interface (Sect.~\ref{sec:library_features}).
    \item \textbf{Reproducibility by construction.} Each dataset records the environment specification, generation seeds, applied wrappers, and pinned package requirements, so the generating environment can be rebuilt and the data regenerated. Earlier offline RL benchmarks store none of this (Sect.~\ref{sec:repro}).
    \item \textbf{A dataset collection.} \numdatasets datasets (\numtrajectories trajectories, \numsteps steps) in six collections, each documented and released with its generation script. This includes a regeneration of D4RL on maintained versions of MuJoCo and Gymnasium (Sect.~\ref{sec:datasets}).
    \item \textbf{Baselines.} Scores for six offline RL algorithms, run with CORL~\citep{corl}, on the regenerated D4RL and MuJoCo datasets (Sect.~\ref{sec:benchmarks}).
\end{itemize}

Minari is maintained by the Farama Foundation alongside Gymnasium, and TorchRL, AgileRL, and d3rlpy already load its datasets natively (Sect.~\ref{sec:integrations}).

\begin{figure}[t]
    \centering
   \begin{pythonsmall}
  import minari
  import gymnasium as gym
  env = gym.make('CartPole-v1')
  env = minari.DataCollector(env, record_infos=True)
  ... # Step through environment as usual
  dataset = env.create_dataset("cartpole/test-v0")\end{pythonsmall}
   \begin{pythonsmall}
  import minari
  dataset = minari.load_dataset('D4RL/door/human-v2', download=True)
  episodes = dataset.sample_episodes(n_episodes=5) # Sample 5 episodes
  env = dataset.recover_environment() # Optionally recover data collection env\end{pythonsmall}
  \caption{Two examples of Minari usage. \textbf{Top:} Creating a Minari dataset using the Gymnasium wrapper. Non-Gymnasium environments are also supported and can be created manually via \lstinline|minari.create_dataset_from_buffers()|. \textbf{Bottom:} Loading an existing dataset (local or remote). In many cases one can instead directly use one of the integrations with existing Offline RL  libraries described in Sect.~\ref{sec:integrations}.}
    \label{fig:code}
\end{figure}

\section{Related work}
Prior work falls into two groups: fixed benchmark releases (D4RL, RL Unplugged) and dataset tooling tied to one framework or domain (RLDS, LeRobot). Table~\ref{tab:comparison} summarizes the differences. Minari is the only entry that lets users create and publish their own datasets, records what is needed to regenerate them, and does not depend on a particular deep learning framework.

\begin{table}[h]
    \centering
    \footnotesize
    \setlength{\tabcolsep}{4pt}
    \resizebox{\textwidth}{!}{%
    \begin{tabular}{@{}lcccccc@{}}
    \toprule
     & \textbf{Standard} & \textbf{User-created} & \textbf{Repro-} & \textbf{Hub} & \textbf{Framework-} & \textbf{Domain} \\
     & \textbf{format} & \textbf{datasets} & \textbf{ducible} & \textbf{sharing} & \textbf{agnostic} & \textbf{scope} \\
    \midrule
    D4RL \citep{d4rl}                 & \checkmark & --         & --         & --         & \checkmark & RL \\
    RL Unplugged \citep{rlunplugged}  & \checkmark & --         & --         & --         & \checkmark & RL \\
    RLDS \citep{rlds}                 & \checkmark & \checkmark & --         & partial    & TF only    & RL \\
    LeRobot \citep{lerobot}           & \checkmark & \checkmark & partial    & \checkmark & PyTorch    & Robotics \\
    \textbf{Minari (ours)}           & \checkmark & \checkmark & \checkmark & \checkmark & \checkmark & General \\
    \bottomrule
    \end{tabular}%
    }
    \caption{Comparison of Minari with related dataset libraries and benchmarks for sequential decision-making. ``Reproducible'' denotes built-in mechanisms to reconstruct the data-generating environment (specs, seeds, wrappers, and software requirements); ``Hub sharing'' denotes one-command publishing/downloading to a community hub.}
    \label{tab:comparison}
\end{table}

\paragraph{RLDS} RLDS \citep{rlds} defines a schema for RL trajectories and a set of operations on them, built on TensorFlow Datasets. Minari's episode-level schema owes a lot to it. The catch is the dependency: using RLDS means bringing in TensorFlow, which most current RL code bases, written in PyTorch or JAX, do not otherwise need.

\paragraph{D4RL} D4RL \citep{d4rl} made offline RL a benchmarkable problem, and most offline RL papers since report results on it. But its datasets were generated with specific, now-unmaintained versions of Gym and mujoco-py, and the generation settings were not stored with the data. As those dependencies age, the datasets become hard to load and impossible to regenerate exactly. D4RL also offers no way for users to create datasets of their own in the same format. Minari regenerates the D4RL tasks on maintained software (Sect.~\ref{sec:d4rl}) and gives users the same tools we used to do it.

\paragraph{RL Unplugged} RL Unplugged \citep{rlunplugged} provides offline datasets for the DeepMind Control Suite \citep{deepmindcontrolsuite}, Atari, and other domains, with more emphasis on hard exploration settings than D4RL. Like D4RL, it is a fixed release. It has no tooling for creating compatible datasets and records no information for regenerating the data.

\paragraph{LeRobot} LeRobot \citep{lerobot} provides models, datasets, and tools for real-world robot learning in PyTorch, with datasets hosted on the HuggingFace Hub. Its format is built around robot data, such as multi-camera video streams and joint states. Minari targets the general case: any Gymnasium space, any environment, simulated or not.

\paragraph{HuggingFace Hub} The Hub hosts most publicly released models and datasets in machine learning and handles versioning, documentation, and access control. It has no notion of RL episodes, spaces, or environments. Minari uses the Hub as its storage backend and adds that structure on top.

\section{Library integrations}
\label{sec:integrations}
Three offline RL libraries load Minari datasets natively, so users can train on any Minari dataset without writing a data loader.

\paragraph{TorchRL} TorchRL \citep{torchrl} is the PyTorch project's RL library, built from modular components for environments, replay buffers, losses, and agents. Minari datasets load directly into its replay buffers.

\paragraph{AgileRL}
AgileRL \citep{agilerl} uses evolutionary hyperparameter optimization and population-based training to tune agent architectures and hyperparameters during training, with the aim of reducing the tuning burden for practitioners.

\paragraph{d3rlpy}
d3rlpy \citep{d3rlpy} is a PyTorch library of offline RL algorithms, including CQL \citep{CQL}, BCQ \citep{BCQ}, and TD3+BC \citep{TD3+BC}, with a short, scikit-learn-style API.

\section{Library features}
\label{sec:library_features}

\subsection{Dataset standards}
Every Minari dataset has an identifier of the form
\begin{verbatim}
    namespace/environment/dataset_name-v(version)
\end{verbatim}
where \verb|namespace| groups related datasets, \verb|environment| names the environment that produced the data, \verb|dataset_name| describes the content, and the version number tracks revisions. Only the dataset name is mandatory. A dataset of human demonstrations on \textit{AdroitHandDoor} could be called \verb|door/human-v0|, or \verb|D4RL/door/human-v0| inside the D4RL namespace. The convention follows Gymnasium environment IDs on purpose, so the identifier alone tells a user what the data is and which revision they have.

Metadata is stored at two levels. Dataset metadata (Appendix~\ref{app:metadata}) records the number of episodes and steps, the observation and action spaces, the environment specification when one exists, a link to the generating code, authors, the algorithm that produced the data, and the pip requirements of the environment. Episode metadata holds reward statistics (minimum, maximum, mean, standard deviation, and sum), which makes it cheap to filter episodes, for example to keep the top 10\% by return for BC-10\%.

The data itself is a sequence of episodes. Each episode holds observations, actions, rewards, termination and truncation flags, and optional info dictionaries. Observations and actions can be any Gymnasium space, including nested \verb|Dict| and \verb|Tuple| spaces and \verb|Text| (Appendix~\ref{app:spaces}), and the episode structure stays the same regardless. Terminations and truncations are kept separate because offline RL algorithms need the distinction: bootstrapping from a time-limit truncation as if it were a terminal state biases value estimates.

Storage is pluggable behind the same API. HDF5 is the default and suits small and medium datasets, while the Arrow/Parquet backend is meant for large datasets and distributed training.

\subsection{Minari software package}
\begin{table}[]
    \centering
    \begin{tabular}{cc}
    \toprule
    \textbf{Command name} & \textbf{Description}\\
    \midrule
     \bashinline{list local} & List Minari datasets in local storage. \\
     \bashinline{list remote} & List Minari datasets in remote storage. \\
     \bashinline{show} & Describe a local or remote dataset, and its environment. \\
     \bashinline{delete} & Delete datasets from local database. \\
     \bashinline{download} & Download Minari datasets from remote server.\\
     \bashinline{upload} & Upload Minari datasets to the specified remote server. \\
     \bashinline{combine} & Combine multiple datasets into a single dataset.\\
     \bottomrule
    \end{tabular}
    \caption{List of available commands in the Minari command line interface, along with their description. To use a command append minari before the command name, e.g. \textit{minari list local}; \textit{-\,-help} can be used for more information.}
    \label{tab:cli}
\end{table}

Minari is a pure Python package installable with \verb|pip install minari|. It covers four tasks: creating, loading, manipulating, and sharing datasets.

\paragraph{Creating datasets.} There are two ways to create a dataset. The \verb|DataCollector| wrapper records every step of a Gymnasium environment while the agent runs, with no other changes to the training or rollout code (Fig.~\ref{fig:code}, top). For data that does not come from a Gymnasium environment, such as logs from another simulator, a real robot, or an existing dataset in another format, \verb|create_dataset_from_buffers()| takes a list of episode buffers and writes them in the Minari format. For pixel observations, Minari can optionally JPEG-encode frames, which shrinks on-disk and download size for datasets such as Atari.

\paragraph{Loading and manipulating datasets.} Loaded datasets support episode sampling, iteration, filtering by any episode-level predicate, and shuffling. Several datasets with compatible spaces can be combined into one.

\paragraph{Sharing datasets.} The \verb|minari upload| command publishes a dataset to the HuggingFace Hub, and \verb|minari download| fetches it, along with its metadata, so another group can reproduce or extend an experiment from the dataset ID alone. One-off operations like these are also available from the command line without writing Python (Table~\ref{tab:cli}).

\subsection{Improving reproducibility in Offline RL}
\label{sec:repro}

RL results are notoriously hard to reproduce \citep{hendersonDeepReinforcementLearning2019}. Offline RL adds a failure mode of its own: a policy is trained on data from one environment and evaluated in another that is supposed to be identical but is not. A changed time limit, a missing wrapper, or a different simulator version is enough to make scores incomparable, and nothing in a bare array of transitions reveals the mismatch. Minari stores three things to prevent it.

\paragraph{Environment and wrappers.} \pythoninline{dataset.recover_environment()} rebuilds the environment that generated the data from its stored specification, including any wrappers that changed its behavior. Evaluation therefore runs on the same environment as data collection by default, not on whatever the current release of the environment happens to be. When the evaluation task should differ from the collection task, for example a goal-conditioned dataset collected with random goals but evaluated on a fixed one, the dataset can also store a separate evaluation environment, returned by \pythoninline{recover_environment(eval_env=True)}.

\paragraph{Seeds.} When the user does not pass a seed to \verb|reset()|, the \verb|DataCollector| draws one and stores it with the episode. Rerunning the same policy (with its own randomness seeded) from the stored seeds regenerates the same episodes.

\paragraph{Software versions.} Physics simulators such as MuJoCo~\citep{mujoco} integrate rigid-body dynamics numerically, so the same policy can produce different trajectories under different simulator versions. Dataset creators can pin pip requirements when creating a dataset, and Minari checks the installed packages against them when the environment is recovered.

\section{Available datasets}
\label{sec:datasets}

We host \numdatasets datasets on the HuggingFace Hub, with \numtrajectories trajectories and \numsteps steps in total (\datasetsize). Table~\ref{tab:collections} breaks them down by collection. They cover discrete and continuous actions and state, grid, pixel, and text observations, and the data comes from planners, trained RL agents, scripted experts, and humans. The script that generated each dataset is public\footnote{\url{https://github.com/Farama-Foundation/minari-dataset-generation-scripts}}, and each dataset has a documentation page with its specification and statistics\footnote{\url{https://minari.farama.org}}. Anyone can therefore regenerate a dataset or produce a variant of it, for instance with a different policy or seed set.

\begin{table}[h]
    \centering
    \footnotesize
    \begin{tabular}{@{}lllll@{}}
    \toprule
    \textbf{Collection} & \textbf{Domain} & \textbf{Observations} & \textbf{\# Datasets} & \textbf{Data source} \\
    \midrule
    D4RL      & Locomotion, manipulation, navigation & State / grid & 31  & Planners, RL, human, IL \\
    MuJoCo    & Continuous control                   & State        & 26  & SAC / TQC (SB3) \\
    BabyAI    & Instruction following                & Grid / text  & 170 & Expert bot policy \\
    Atari     & Arcade games                         & Pixels       & 57  & PPO (CleanRL) \\
    MetaWorld & Robotic manipulation                 & State        & 50  & Scripted expert \\
    WebAgents & Web navigation                       & Text         & 4   & Human (WebLINX) \\
    \bottomrule
    \end{tabular}
    \caption{Overview of the dataset collections shipped with Minari. Counts are approximate and grow over time; per-dataset statistics (episodes, steps, returns) are available in the documentation and in each dataset's metadata. Together these collections span discrete and continuous actions and state-, grid-, pixel-, and text-based observations.}
    \label{tab:collections}
\end{table}

\subsection{D4RL}
\label{sec:d4rl}
\begin{figure}[t!]
    \centering
    \includegraphics[width=0.18\linewidth]{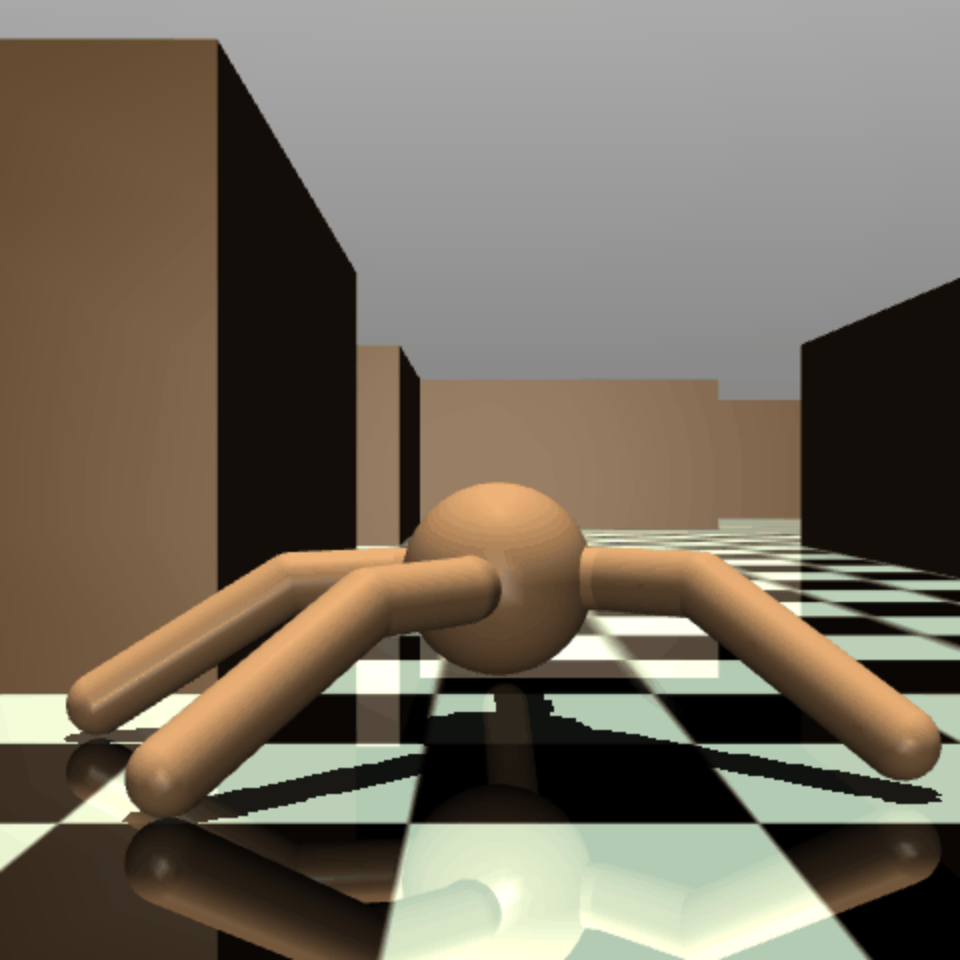}
    \includegraphics[width=0.18\linewidth]{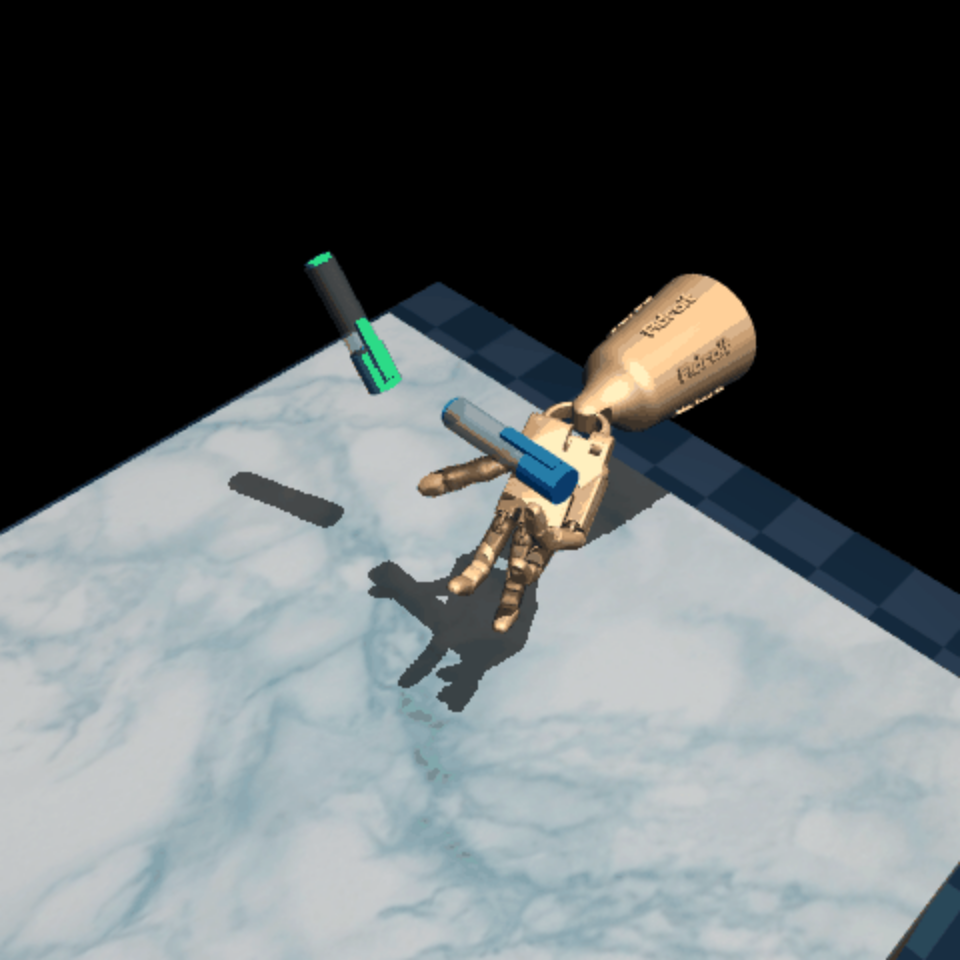}
    \includegraphics[width=0.18\linewidth]{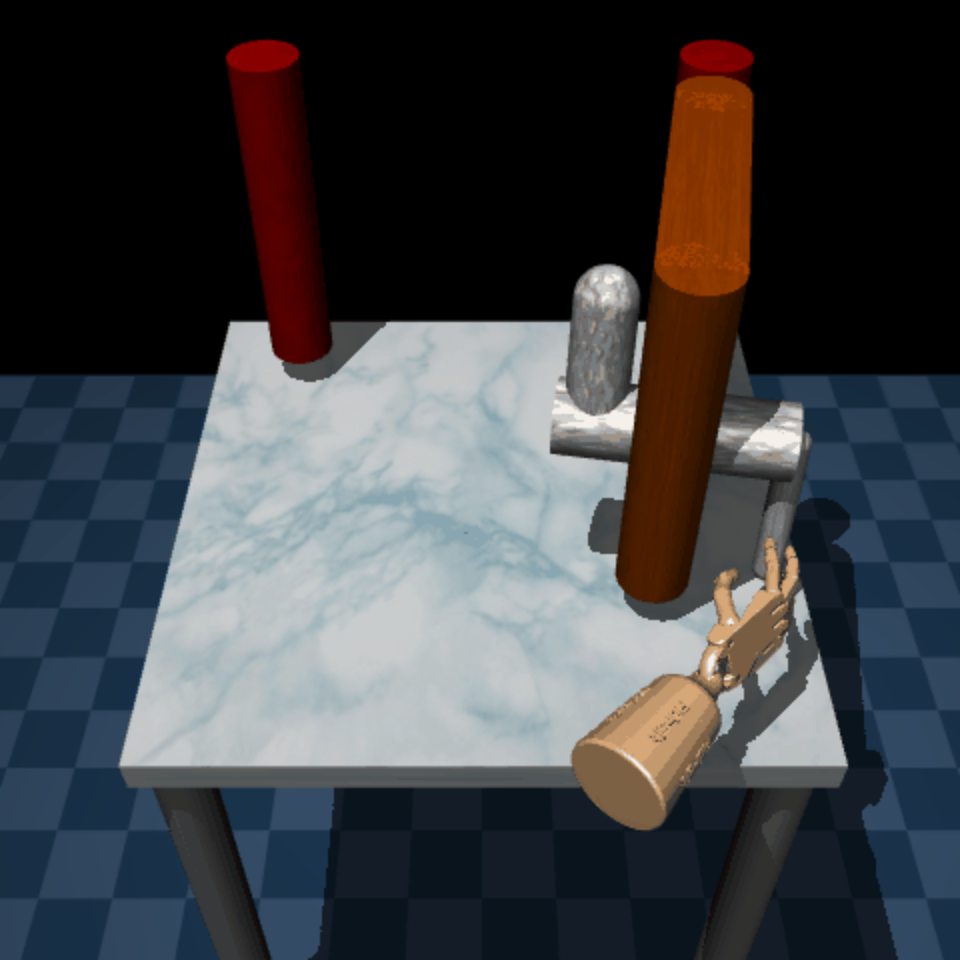}
    \includegraphics[width=0.18\linewidth]{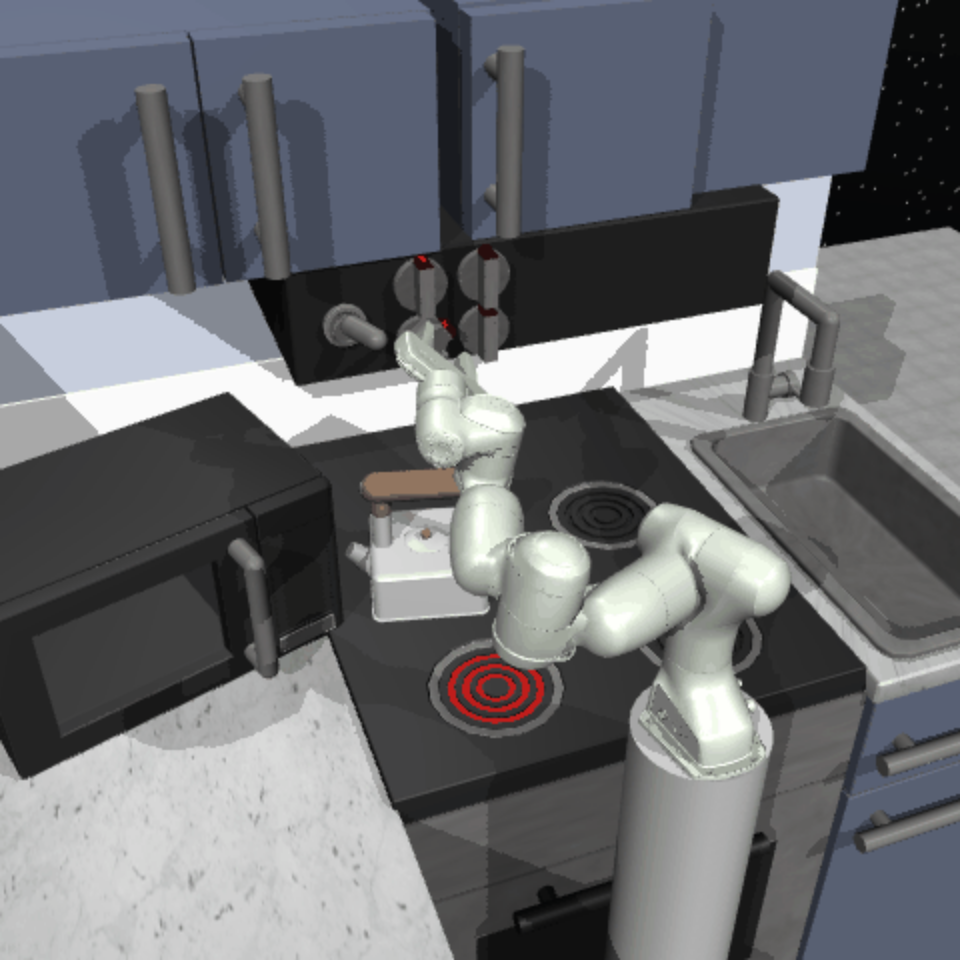}
    \includegraphics[width=0.18\linewidth]{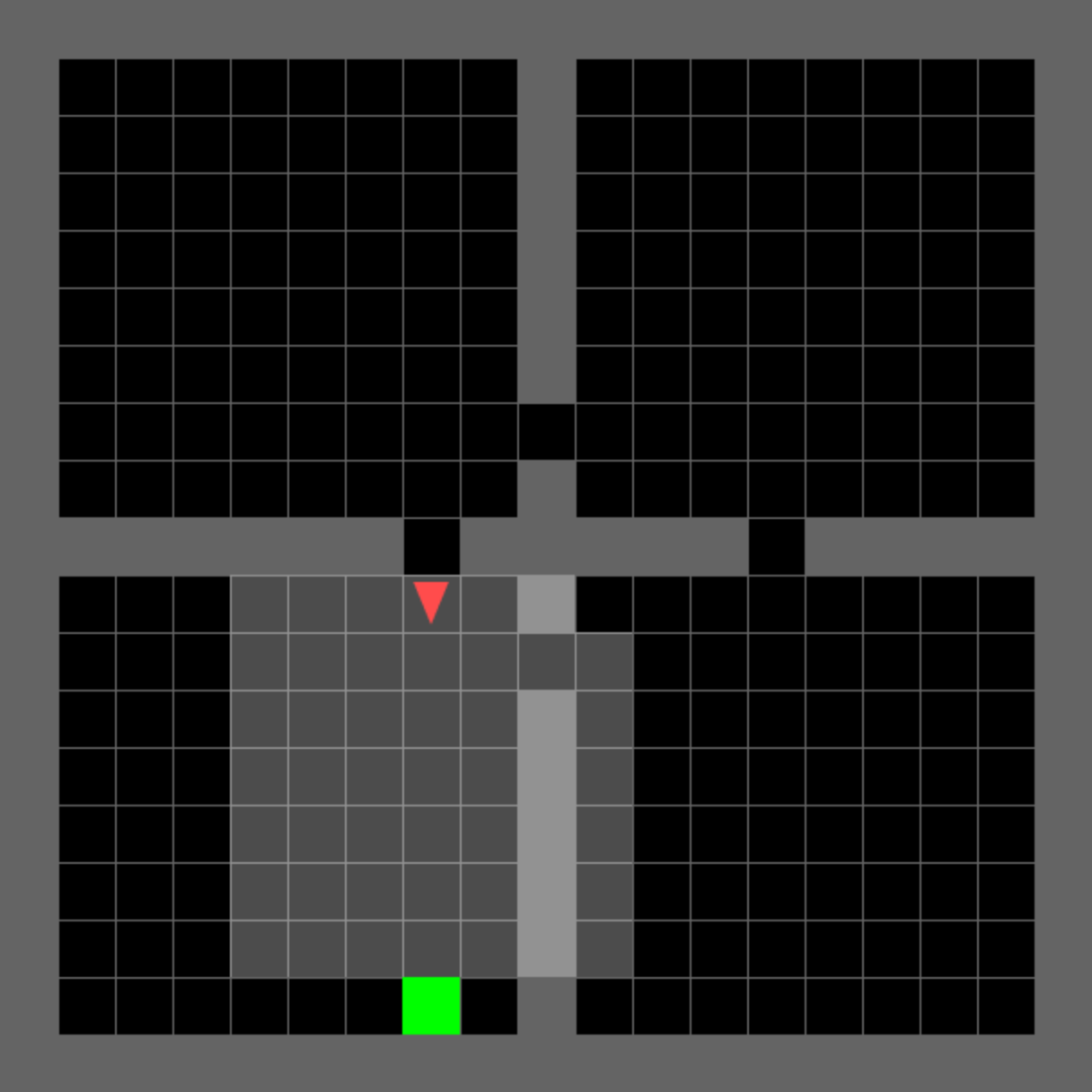}
    \caption{Representative D4RL environments (left to right): \textsc{Ant Maze}, \textsc{Pen}, \textsc{Door}, \textsc{Kitchen}, and \textsc{Minigrid}. \textsc{Point Maze}, \textsc{Hammer}, and \textsc{Relocate} are also in the collection but not shown.}
    \label{fig:dr4l}
\end{figure}
\begin{table}[]
\footnotesize
\centering
\begin{tabular}{@{}lllllll@{}}
\toprule
                             & BC    & BC-10\% & TD3+BC & AWAC  & CQL   & IQL   \\ \midrule
pointmaze/umaze-v2           & 53.8  & 55.1    & 71.9   & 52.3  & 85.5  & 52.4  \\
pointmaze/medium-v2          & 78.8  & 59.3    & 105.8  & 85.3  & 80.7  & 38.6  \\
pointmaze/large-v2           & 42.3  & 36.9    & 80.1   & 111.8 & 63.3  & 43.3  \\ \midrule
\textit{pointmaze average}   & 58.3  & 50.4    & 85.9   & 83.1  & 76.5  & 44.8  \\ \midrule
antmaze/umaze-v2             & 97.7  & 92.0    & 97.7   & 100.0 & 46.3  & 87.7  \\
antmaze/umaze-diverse-v2     & 95.3  & 89.7    & 96.7   & 96.7  & 79.0  & 87.7  \\
antmaze/medium-play-v2       & 36.7  & 34.0    & 59.3   & 70.0  & 33.0  & 41.3  \\
antmaze/medium-diverse-v2    & 16.3  & 11.0    & 23.3   & 56.7  & 5.7   & 21.3  \\
antmaze/large-play-v2        & 20.0  & 22.7    & 32.7   & 86.7  & 20.7  & 29.0  \\
antmaze/large-diverse-v2     & 16.0  & 14.7    & 23.3   & 26.7  & 6.7   & 24.3  \\ \midrule
\textit{antmaze average}     & 47.0  & 44.0    & 55.5   & 72.8  & 31.9  & 48.6  \\ \midrule
pen/human-v2                 & 56.9  & 17.5    & 0.4    & 77.3  & 35.4  & 64.0  \\
pen/cloned-v2                & 68.0  & 42.0    & 36.6   & 77.8  & 14.3  & 79.6  \\
pen/expert-v2                & 110.4 & 62.2    & 114.7  & 118.7 & 18.8  & 117.1 \\ \midrule
\textit{pen average}         & 78.4  & 40.6    & 50.6   & 91.3  & 22.8  & 86.9  \\ \midrule
door/human-v2                & 6.0   & 0.4     & 0.5    & 13.9  & 8.0   & 9.8   \\
door/cloned-v2               & 0.3   & 5.2     & 0.5    & 5.7   & 0.5   & 5.4   \\
door/expert-v2               & 104.3 & 105.2   & 2.8    & 105.1 & 85.7  & 106.1 \\ \midrule
\textit{door average}        & 36.9  & 36.9    & 1.3    & 41.6  & 31.4  & 40.4  \\ \midrule
hammer/human-v2              & 15.4  & 10.4    & 6.6    & 27.6  & 0.3   & 11.6  \\
hammer/cloned-v2             & 14.8  & 17.4    & 2.1    & 6.5   & 0.3   & 27.6  \\
hammer/expert-v2             & 131.5 & 132.9   & 102.5  & 132.6 & 0.3   & 133.4 \\ \midrule
\textit{hammer average}      & 53.9  & 53.6    & 37.1   & 55.6  & 0.3   & 57.5  \\ \midrule
relocate/human-v2            & 1.5   & 0.2     & 0.2    & 1.2   & 0.2   & 0.7   \\
relocate/cloned-v2           & 0.2   & 0.2     & 0.3    & 0.2   & 0.3   & 0.7   \\
relocate/expert-v2           & 106.7 & 98.4    & 1.2    & 107.2 & 0.3   & 106.7 \\ \midrule
\textit{relocate average}    & 36.1  & 32.9    & 0.6    & 36.2  & 0.3   & 36.0  \\ \bottomrule
\end{tabular}
\caption{Normalized scores on the regenerated D4RL navigation and manipulation datasets, using the CORL~\citep{corl} implementations. Each cell is the mean over 3 seeds of the best checkpoint, evaluated with each algorithm's CORL evaluation settings (10--100 episodes); italic rows are group averages. Scores are $100\,(R-R_{\min})/(R_{\max}-R_{\min})$ with the random and expert reference returns stored in each dataset's metadata, as in D4RL~\citep{d4rl}. Unlike D4RL's maze2d, \textsc{Point Maze} is goal-conditioned and uses different reference returns, so these rows are not comparable to published maze2d results.}
\label{tab:benchmarks_d4rl}
\end{table}
D4RL \citep{d4rl} is still the default offline RL benchmark, but its datasets were generated with mujoco-py and versions of Gym that are no longer maintained. We regenerated its tasks on Gymnasium, Gymnasium-Robotics, and the official MuJoCo bindings, following the original data-generation procedures. The regenerated data is not bit-identical to the original, since the simulators and environment implementations changed. In exchange, every dataset can now be rebuilt from its stored seeds and script. The collection contains (Fig.~\ref{fig:dr4l}):
\begin{itemize}
    \item \textbf{Ant Maze} (6 datasets). An 8-DoF quadruped follows waypoints from a Q-value iteration planner through mazes of three sizes, with a success rate above 80\%.
    \item \textbf{Point Maze} (8 datasets). A force-actuated ball moves to random goals, driven by a Q-value iteration planner and a PD controller, in 4 mazes with sparse and dense rewards.
    \item \textbf{Adroit} (12 datasets). A 24-DoF simulated hand performs four dexterous tasks from \citet{d4rl-pen}: rotating a pen to a target orientation, opening a door, hammering a nail, and relocating a ball. As in D4RL, each task has three datasets: the original human demonstrations, rollouts of a trained RL expert (\textit{expert}), and rollouts of a policy trained by imitation on a mix of the two (\textit{cloned}).
    \item \textbf{Kitchen} (3 datasets). A Franka arm manipulates kitchen objects, using data from \citet{relaypolicylearning}. The target is four subtasks: open the microwave, move the kettle, flip the light switch, and slide open the cabinet. Following D4RL, \textit{complete} contains all four in sequence, \textit{mixed} completes them alongside other subtasks, and \textit{partial} never completes all four.
    \item \textbf{Minigrid Four Rooms} (2 datasets). A grid agent navigates to a goal in the Minigrid \citep{minigrid} four-rooms layout, under a random policy and an optimal value-iteration policy.
\end{itemize}

\subsection{MuJoCo}
\begin{figure}[t!]
    \centering
    \includegraphics[width=0.18\linewidth]{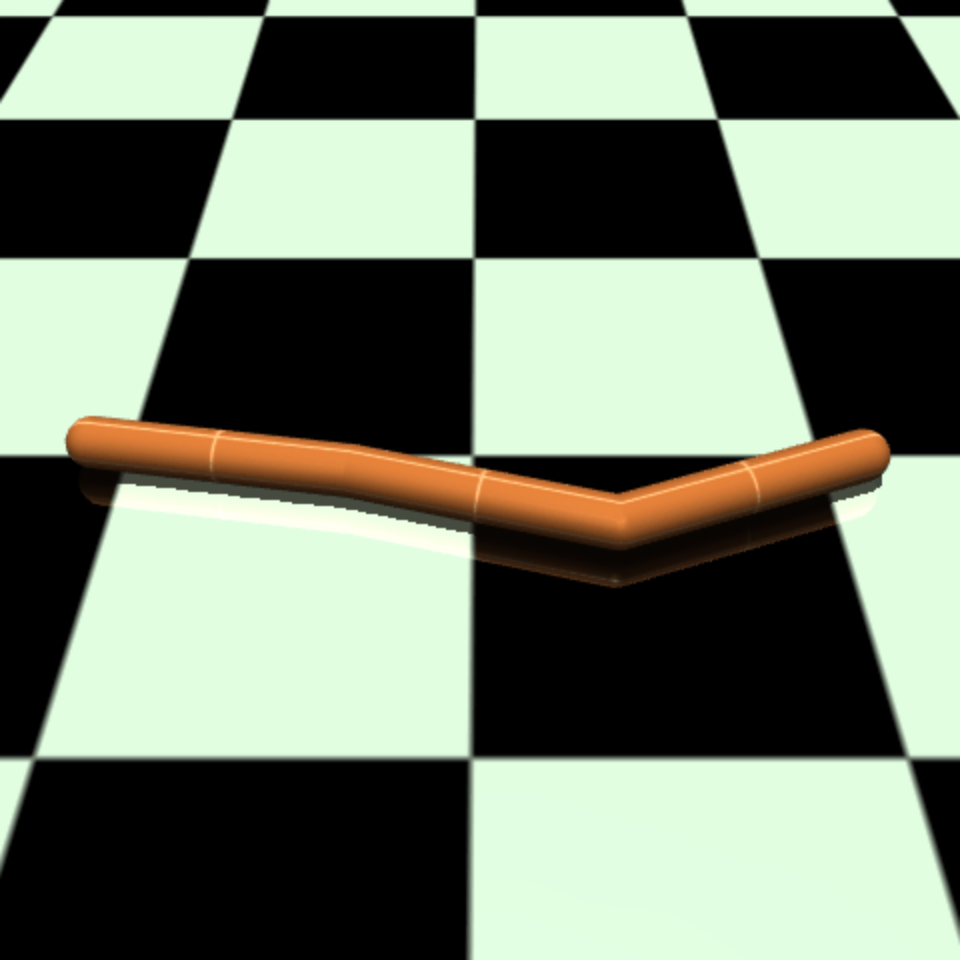}
    \includegraphics[width=0.18\linewidth]{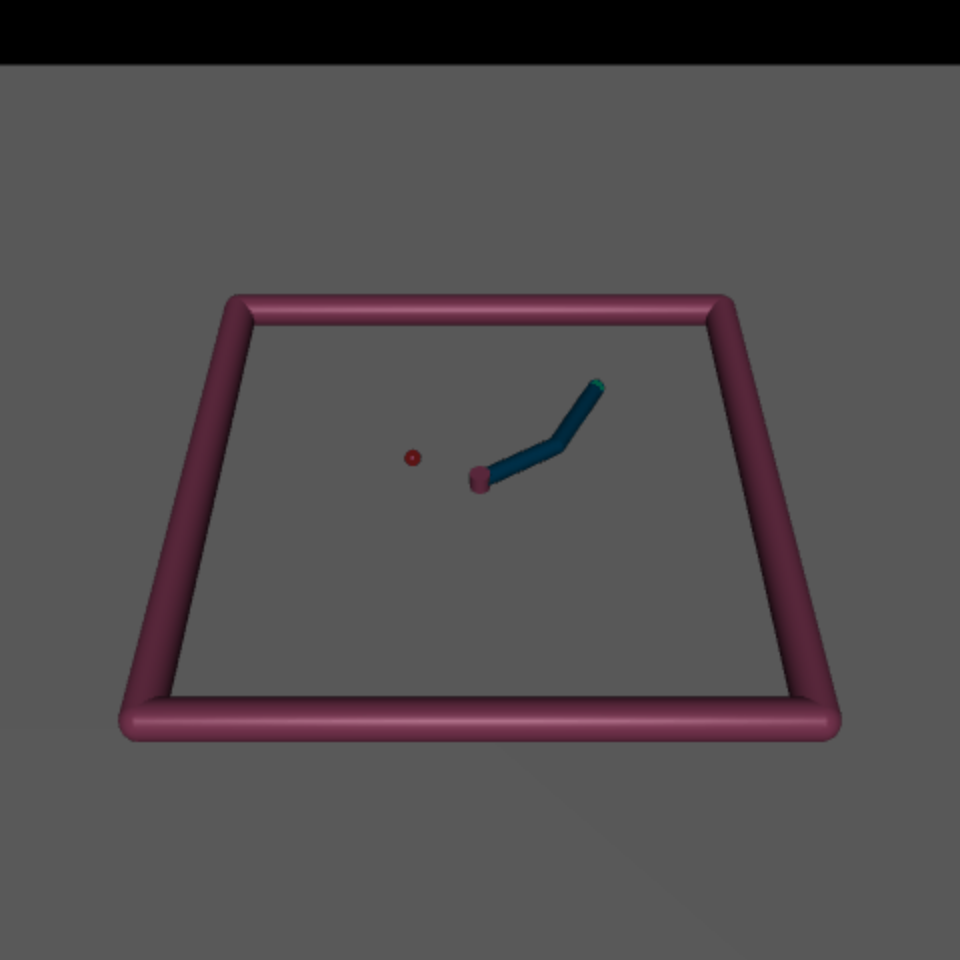}
    \includegraphics[width=0.18\linewidth]{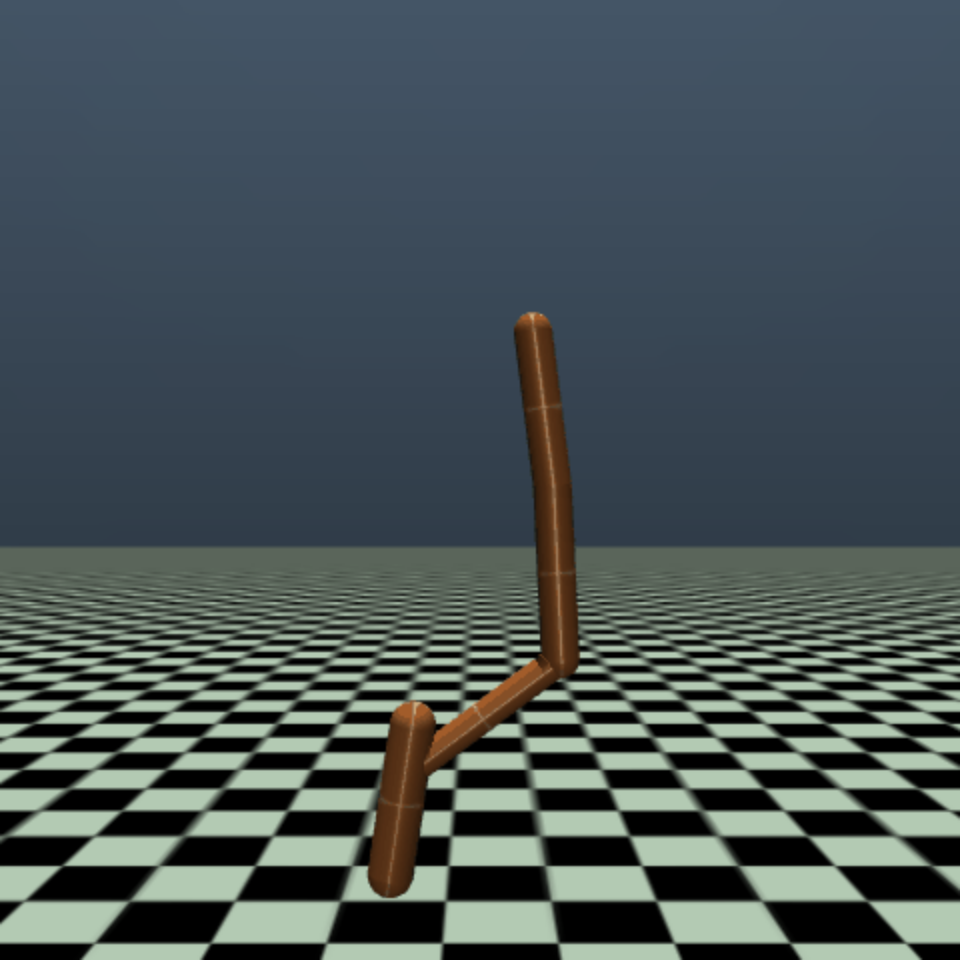}
    \includegraphics[width=0.18\linewidth]{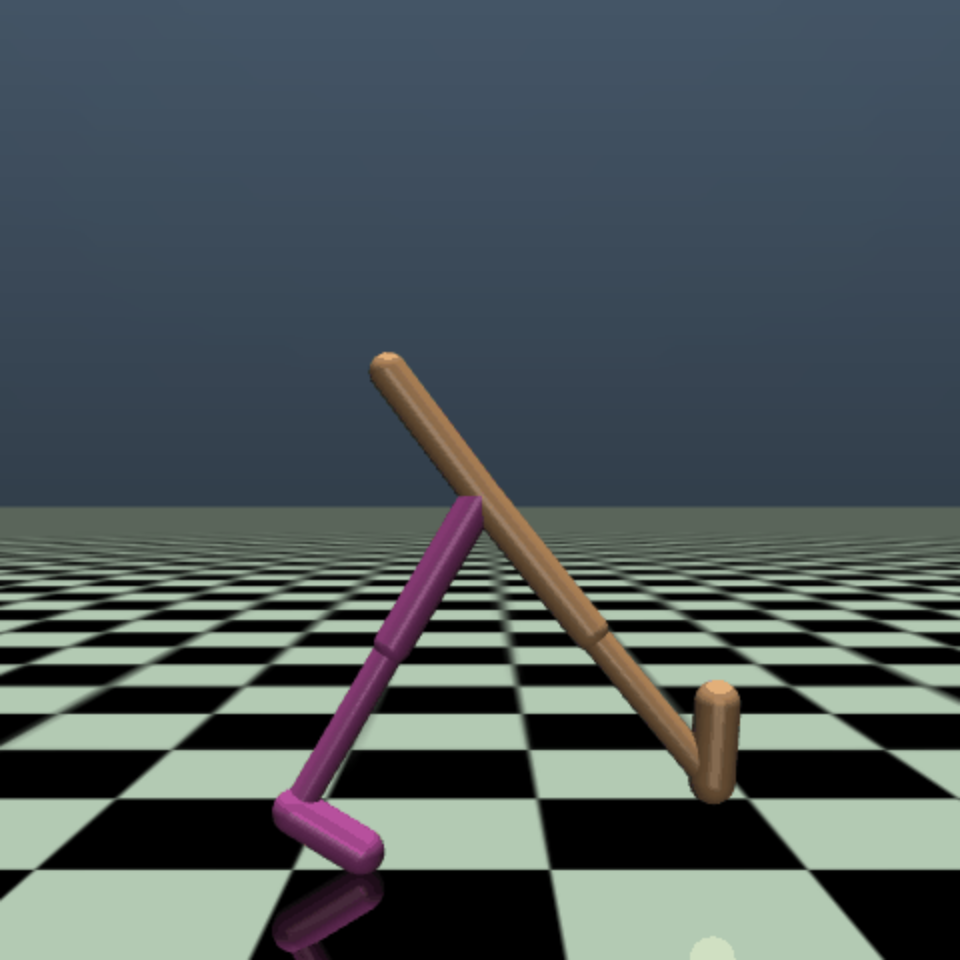}
    \includegraphics[width=0.18\linewidth]{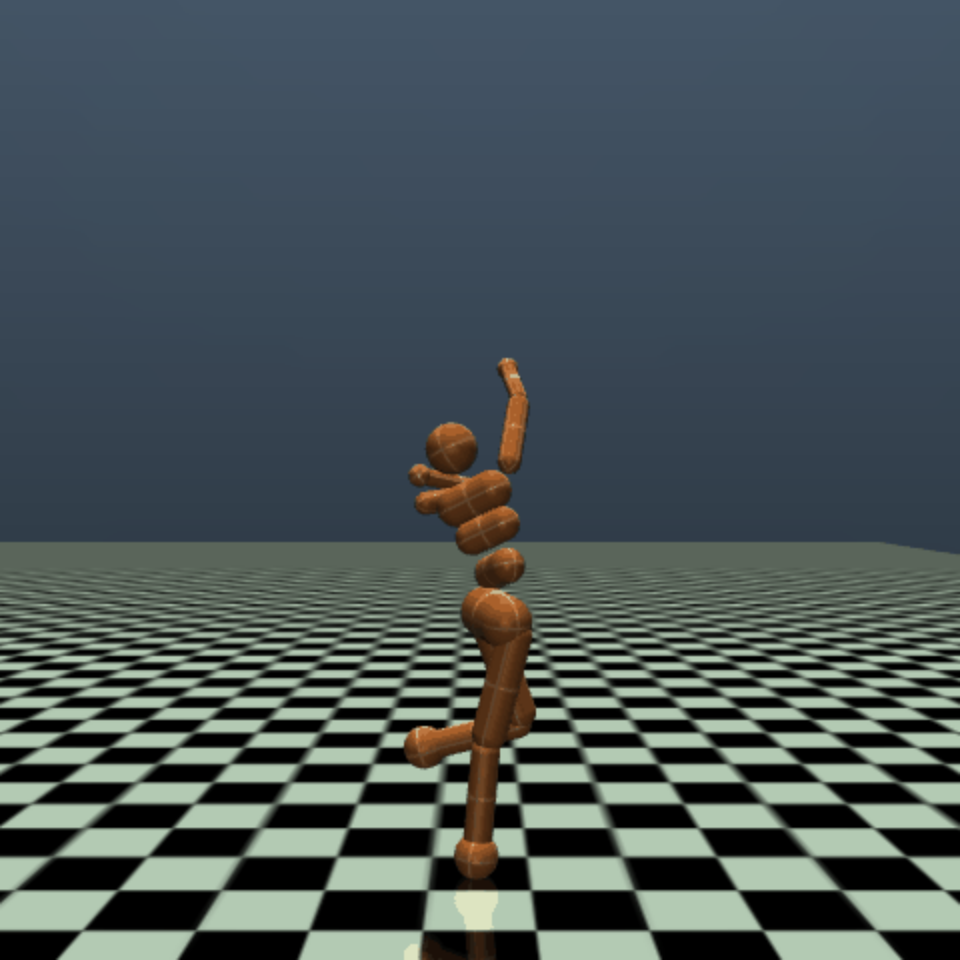}
    \caption{MuJoCo datasets: \textsc{Swimmer}, \textsc{Reacher}, \textsc{Hopper}, \textsc{Walker2d} and \textsc{Humanoid} environments (left to right).}
    \label{fig:mujoco}
\end{figure}
\begin{table}[]
\footnotesize
\centering
\begin{tabular}{@{}lllllll@{}}
\toprule
                             & BC    & BC-10\% & TD3+BC & AWAC  & CQL   & IQL   \\ \midrule
halfcheetah/simple-v0        & 46.0  & 45.7    & 48.8   & 49.5  & 47.8  & 49.1  \\
halfcheetah/medium-v0        & 95.2  & 94.3    & 87.3   & 95.9  & 88.8  & 93.8  \\
halfcheetah/expert-v0        & 105.6 & 79.5    & 50.0   & 108.5 & 65.2  & 107.1 \\ \midrule
\textit{halfcheetah average} & 82.3  & 73.2    & 62.0   & 84.6  & 67.3  & 83.4  \\ \midrule
hopper/simple-v0             & 83.4  & 83.6    & 83.6   & 83.9  & 82.6  & 83.7  \\
hopper/medium-v0             & 94.5  & 94.7    & 94.5   & 96.4  & 94.0  & 94.7  \\
hopper/expert-v0             & 108.4 & 109.1   & 107.9  & 108.1 & 107.1 & 106.1 \\ \midrule
\textit{hopper average}      & 95.4  & 95.8    & 95.4   & 96.1  & 94.5  & 94.8  \\ \midrule
walker2d/simple-v0           & 61.9  & 62.4    & 63.8   & 62.1  & 62.0  & 64.4  \\
walker2d/medium-v0           & 91.0  & 91.5    & 91.5   & 91.2  & 91.8  & 91.5  \\
walker2d/expert-v0           & 101.9 & 102.1   & 101.4  & 29.4  & 100.3 & 102.3 \\ \midrule
\textit{walker2d average}    & 85.0  & 85.3    & 85.6   & 60.9  & 84.7  & 86.1  \\ \bottomrule
\end{tabular}
\caption{Normalized scores on the Minari MuJoCo locomotion datasets, using the CORL~\citep{corl} implementations. Each cell is the mean over 3 seeds of the best checkpoint over $10^6$ gradient steps, evaluated for 10 episodes; italic rows average the three skill tiers. $R_{\min}$ is the mean return of a random policy (50 episodes) and $R_{\max}$ the mean return of the corresponding \textit{expert} dataset. Minari's SAC/TQC experts are stronger than D4RL's reference experts ($1.34\times$ on \textsc{HalfCheetah}, $1.49\times$ on \textsc{Walker2d}), so a score of 100 here means a higher return than in D4RL and the two are not directly comparable.}
\label{tab:benchmarks_mujoco}
\end{table}
The MuJoCo collection covers every MuJoCo \citep{mujoco} environment in Gymnasium \citep{andreasthesis}, on current environment versions (Fig.~\ref{fig:mujoco}). Unlike the D4RL locomotion datasets, it is not a regeneration of an earlier release. We trained agents with Stable Baselines3 \citep{sb3}, using SAC \citep{sac} everywhere except \textsc{Humanoid}, where we used TQC \citep{tqc}. Checkpoints saved at different performance plateaus give datasets at several skill levels, since offline RL methods are expected to work from suboptimal data as well as from experts.

The balancing tasks (\textsc{InvertedPendulum}, \textsc{InvertedDoublePendulum}) have an expert dataset, which never drops the pole, and a medium one, which keeps it up for about 100 steps. The arm tasks (\textsc{Reacher}, \textsc{Pusher}) and \textsc{Swimmer} also have expert and medium datasets. The planar locomotion tasks (\textsc{HalfCheetah}, \textsc{Hopper}, \textsc{Walker2d}) add a third, \textit{simple}, tier below medium. So do \textsc{Ant}, whose expert gallops fast but sometimes falls, while the medium agent walks quickly and stably and the simple one moves slowly; and \textsc{Humanoid}, which ranges from running (expert) to brisk walking (medium) to a slow sliding gait (simple).

\subsection{BabyAI}
\begin{figure}[t!]
    \centering
    \includegraphics[width=0.18\linewidth]{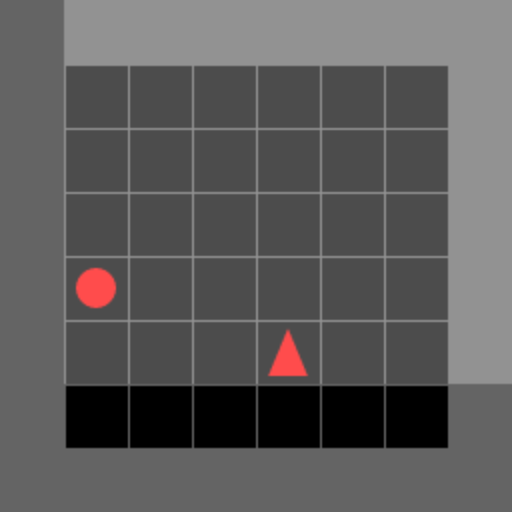}
    \includegraphics[width=0.18\linewidth]{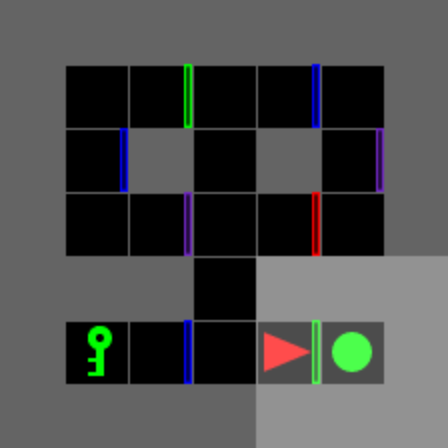}
    \includegraphics[width=0.18\linewidth]{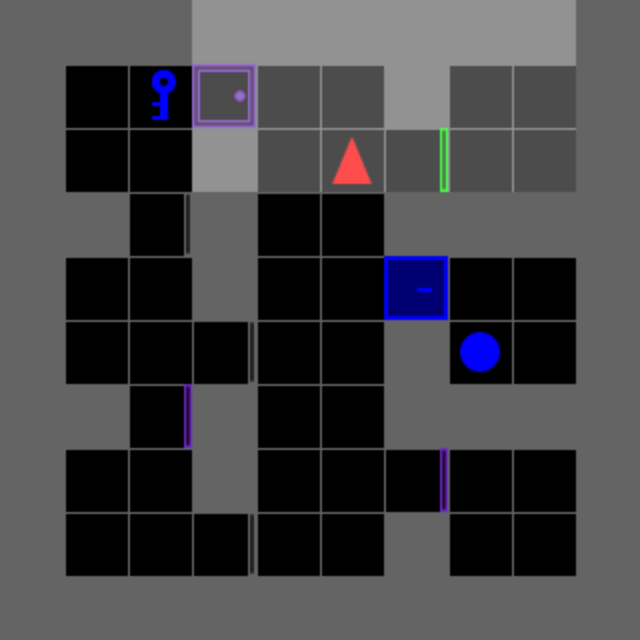}
    \includegraphics[width=0.384\linewidth]{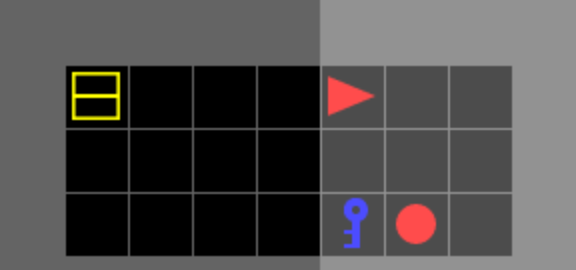}
    \caption{BabyAI datasets: \textsc{GoToRedBall}, \textsc{KeyCorridor} (two variants), and \textsc{MoveTwoAcross} tasks (left to right).}
    \label{fig:babyai}
\end{figure}
BabyAI \citep{babyai} is a set of grid-world tasks for grounded language learning. In each episode the agent receives a natural-language instruction, such as ``put the red ball next to the blue box'', and must carry it out from partial observations. Tasks range from going to a single object to long instructions that chain several subgoals.

We generated 170 datasets covering every BabyAI level (Fig.~\ref{fig:babyai}), from \textsc{GoToRedBall} up to \textsc{BossLevel}. Each level has two versions: one with the standard agent-centric partial view and one with the full grid as observation, so the same task can be studied with and without partial observability. Trajectories come from the BabyAI bot, a deterministic expert that solves every level, and each dataset holds 1,000 episodes. Because the instruction is part of the observation (a \verb|Text| space inside a \verb|Dict|), these datasets also exercise Minari's support for structured, non-numeric observations.

\subsection{Atari}
\begin{figure}[t!]
    \centering
    \includegraphics[width=0.18\linewidth]{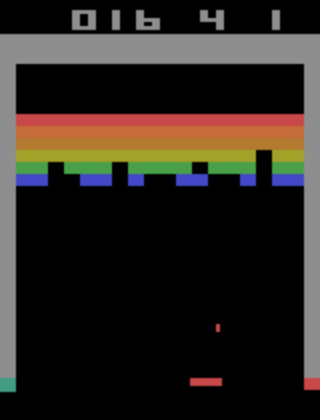}
    \includegraphics[width=0.18\linewidth]{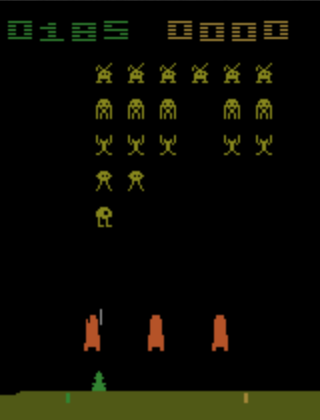}
    \includegraphics[width=0.18\linewidth]{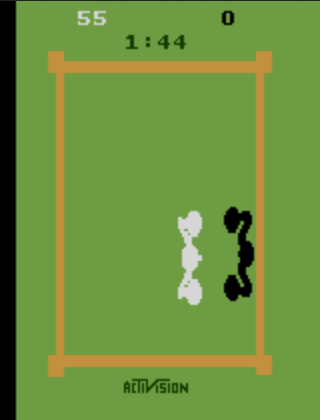}
    \includegraphics[width=0.18\linewidth]{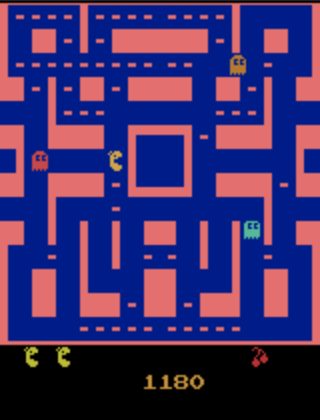}
    \includegraphics[width=0.18\linewidth]{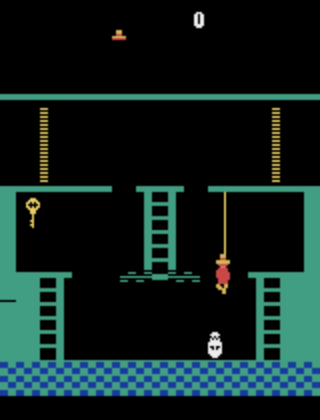}
    \caption{ALE datasets: \textsc{Breakout}, \textsc{Space Invaders}, \textsc{Boxing}, \textsc{Ms PacMan} and \textsc{Montezuma's Revenge} environments (left to right).}
    \label{fig:atari}
\end{figure}
We recorded 10 episodes in each of the 57 Atari 2600 games of the Arcade Learning Environment \citep{ALE} (Fig.~\ref{fig:atari}), using pretrained CleanRL \citep{cleanrl} agents trained with PPO \citep{ppo} and an IMPALA network \citep{impala}. The agents act on the usual preprocessed input (grayscale, resized, frame-stacked), but the datasets store the raw RGB frames, so users can choose their own preprocessing. The games range from shooters (\textsc{Space Invaders}) and maze games (\textsc{Ms. Pac-Man}) to sports (\textsc{Boxing}) and hard-exploration games (\textsc{Montezuma's Revenge}). This collection is where Minari's optional JPEG encoding of pixel observations matters most for storage and download size.

\subsection{MetaWorld}
MetaWorld \citep{metaworld} has 50 manipulation tasks for a simulated Sawyer arm, built for multi-task and meta-RL: reaching, pushing, pick-and-place, tool use, opening doors and drawers, pressing buttons, and more. We recorded 50 episodes per task with the scripted expert policies that ship with MetaWorld, which solve nearly every episode. Observations are the full state: hand position, gripper opening, object poses, and goal position. The Adroit datasets offer few tasks with a high-dimensional hand; MetaWorld offers many tasks with a simple gripper and binary success signals, which suits multi-task offline RL and studies of transfer between tasks.

\subsection{WebAgents}
\label{sec:webagents}
We converted WebLINX \citep{weblinx} to the Minari format. It contains 969 successful demonstrations of web navigation tasks, about 24K interaction steps in total, covering form filling, information lookup, and multi-step workflows on real websites. Observations and actions are text and there is no reward. We include it to show that the format is not limited to simulated control: any sequential decision-making data that fits Gymnasium spaces fits in Minari, including data for training language-model agents.

\section{Benchmarks}
\label{sec:benchmarks}

Minari is infrastructure, and benchmarking algorithms is better left to the training libraries built on it (Sect.~\ref{sec:integrations}). Still, a regenerated dataset is only useful if standard algorithms learn from it, so we checked. We trained six offline RL algorithms, BC, BC-10\%, TD3+BC~\citep{TD3+BC}, AWAC, CQL~\citep{CQL}, and IQL, on the regenerated D4RL datasets and on the MuJoCo collection, using the CORL~\citep{corl} implementations and their tuned hyperparameters unchanged. Tables~\ref{tab:benchmarks_d4rl} and~\ref{tab:benchmarks_mujoco} report normalized scores.

We report the best checkpoint over training, not the last. Off-policy actor-critic methods, TD3+BC, AWAC, and CQL in particular, can diverge late in training on narrow datasets, and the final checkpoint then understates what the method reached. CORL reports both for the same reason~\citep{corl}. Best-checkpoint scores are optimistic, so they should be compared with CORL's best-checkpoint numbers and not with last-checkpoint results from the original papers.

\paragraph{What the results show.} The regenerated D4RL datasets keep the difficulty structure of the originals. Expert Adroit data is easy for imitation-style methods (BC, AWAC, and IQL all exceed 100 on every expert dataset), while the human and cloned datasets for \textit{door}, \textit{hammer}, and \textit{relocate} stay hard, with no method above 27.6. IQL is the most consistent method across both tables and lands close to CORL's published scores where a direct comparison exists (87.7 vs.\ 87.5 on \textit{antmaze/umaze}, 106.1 vs.\ 106.7 on \textit{door/expert}). Two departures from published results are worth stating plainly. First, BC and AWAC score higher on the regenerated Ant Maze datasets than on the originals, especially on the large mazes. The regenerated data comes from a different maze implementation and planner rollouts, so exact agreement is not expected, but users comparing against older Ant Maze numbers should be aware of the gap. Second, CQL trails the other methods on Ant Maze and collapses on \textit{hammer} and \textit{relocate}, and TD3+BC fails on \textit{door} and \textit{relocate}. CQL's Ant Maze results are known to be hard to reproduce, and in CORL's configuration it is the only one of the six methods trained without state normalization. We report CORL's settings as-is and have not tuned them for the new datasets.

On the MuJoCo collection (Table~\ref{tab:benchmarks_mujoco}), all methods recover most of the expert's return on \textsc{Hopper} and \textsc{Walker2d}, and the datasets separate methods mainly at the expert tier: TD3+BC and CQL lose ground on \textit{halfcheetah/expert}, and AWAC falls to 29.4 on \textit{walker2d/expert}.

These numbers are sanity checks and a starting point, not a leaderboard. The point is that the datasets work with a widely used training library without modification, and that anyone can now rerun these experiments on data they can also regenerate.

\section{Maintenance, hosting, and licensing}
\label{sec:maintenance}
Minari is maintained by the Farama Foundation, the non-profit that also maintains Gymnasium and PettingZoo, which gives the project support beyond the lifespan of a single research grant. The library is MIT-licensed, distributed on PyPI, and developed in the open on GitHub. Datasets are versioned and hosted on the HuggingFace Hub, which provides free, persistent public access, and their generation scripts live in a separate public repository so each dataset can be regenerated from scratch. Converted third-party datasets such as WebLINX keep their original licenses, and authorship is recorded in each dataset's metadata.

\section{Conclusion}
Minari gives offline RL what Gymnasium gave online RL: one interface that datasets and training code can both target. A dataset recorded with a Gymnasium wrapper today can be published with one command, loaded by TorchRL, AgileRL, or d3rlpy with one line, and rebuilt years later from the environment specification, seeds, and package versions it stores. The \numdatasets datasets we host (\numtrajectories trajectories, \numsteps steps) include a regenerated D4RL that runs on maintained software, along with collections for MuJoCo, BabyAI, Atari, MetaWorld, and web navigation.

Minari has limits. It stores data as episodes and reads it in batches of episodes, which fits poorly with lifelong or open-ended tasks that have no clear episode boundaries. Its general-purpose storage can also be slower than a data loader hand-tuned for one training pipeline. And reproducibility is only as good as what the dataset creator records: Minari can store seeds and pinned requirements, but it cannot make a nondeterministic simulator deterministic.

The collection is meant to grow through contributions. Any group that records data with Minari can publish it on the Hub, and every compatible training library can use it right away.

\section*{Acknowledgments}
We thank every contributor to Minari; the full list is at \url{https://github.com/Farama-Foundation/Minari/graphs/contributors}.

\bibliographystyle{plainnat}
\bibliography{refs}

\appendix

\section{Dataset metadata}
\label{app:metadata}

Metadata can be attached to a dataset and to each of its episodes. Table~\ref{tab:metadata} lists the default dataset-level fields, which the user fills in at creation time.

\begin{table}[H]
    \centering
    \small
    \begin{tabular}{@{}llp{0.5\linewidth}@{}}
    \toprule
        Attribute & Type & Description \\ \midrule
        \pythoninline{dataset_id} & \pythoninline{str} & Identifier of the Minari dataset. \\
        \pythoninline{total_episodes} & \pythoninline{int} & Number of episodes in the Minari dataset. \\
        \pythoninline{total_steps} & \pythoninline{int} & Number of steps in the Minari dataset. \\
        \pythoninline{action_space} & \pythoninline{Space} & Gymnasium action space describing actions in dataset. \\
        \pythoninline{observation_space} & \pythoninline{Space} & Gymnasium observation space describing observations in dataset. \\
        \pythoninline{env_spec} & \pythoninline{str} & JSON string of the Gymnasium environment spec. \\
        \pythoninline{code_permalink} & \pythoninline{str} & Link to a repository with the code used to generate the dataset. \\
        \pythoninline{author} & \pythoninline{set of str} & Name of the authors that created the dataset. \\
        \pythoninline{author_email} & \pythoninline{set of str} & Email of the authors that created the dataset. \\
        \pythoninline{algorithm_name} & \pythoninline{str} & Name of the expert policy used to create the dataset. \\
        \pythoninline{minari_version} & \pythoninline{str} & Version of Minari that generated the dataset. \\
        \pythoninline{requirements} & \pythoninline{list of str} & List of requirements in pip-style to load the environment. \\ \bottomrule
    \end{tabular}
    \caption{List of metadata attributes of a Minari dataset.}
    \label{tab:metadata}
\end{table}

Only \pythoninline{dataset_id}, \pythoninline{total_episodes}, and \pythoninline{total_steps} are mandatory, and the last two are computed automatically. Reward statistics for each episode are also computed automatically, and users can add their own episode metadata through a callback.

\section{Supported spaces}
\label{app:spaces}

Observations and actions can use any of the following Gymnasium spaces, nested arbitrarily through \pythoninline{Dict} and \pythoninline{Tuple}:

\begin{table}[H]
    \centering
    \begin{tabular}{lll}
        \toprule
        Space & Description & \\ \midrule
        \pythoninline{Box} & An n-dimensional continuous space. & \\
        \pythoninline{Discrete} & Describes a discrete space with an observation in $\{0, 1, ..., n-1\}$. & \\
        \pythoninline{MultiDiscrete} & Represents the cartesian product of arbitrary Discrete spaces. & \\
        \pythoninline{MultiBinary} & A binary space. The elements are binary arrays. & \\
        \pythoninline{Tuple} & Represents a tuple of spaces. & \\
        \pythoninline{Dict} & Represents a dictionary of spaces. & \\
        \pythoninline{Text} & The elements of this space are bounded strings from a charset. & \\ \bottomrule
    \end{tabular}
\end{table}

\section{Episode data structure}

A loaded dataset is a \pythoninline{MinariDataset}, which supports iterating over and sampling episodes. Each episode is returned as an \pythoninline{EpisodeData} object. For example, the following code samples 10 episodes at random from \pythoninline{D4RL/door/human-v2}:

\begin{python}
import minari
dataset = minari.load_dataset("D4RL/door/human-v2", download=True)
sampled_episodes = dataset.sample_episodes(10)
\end{python}

\pythoninline{sampled_episodes} is a list of 10 \pythoninline{EpisodeData} objects with the following fields:

\begin{table}[H]
    \centering
    \small
    \begin{tabular}{@{}llp{0.45\linewidth}@{}}
    \toprule
        Field & Type & Description \\ \midrule
        \pythoninline{id} & \pythoninline{int} & ID of the episode. \\
        \pythoninline{observations} & \pythoninline{np.ndarray}, \pythoninline{list}, \pythoninline{tuple}, \pythoninline{dict} & Stacked observations for each step. \\
        \pythoninline{actions} & \pythoninline{np.ndarray}, \pythoninline{list}, \pythoninline{tuple}, \pythoninline{dict} & Stacked actions for each step. \\
        \pythoninline{rewards} & \pythoninline{np.ndarray} & Rewards for each step. \\
        \pythoninline{terminations} & \pythoninline{np.ndarray} & Terminations for each step. \\
        \pythoninline{truncations} & \pythoninline{np.ndarray} & Truncations for each step. \\
        \pythoninline{infos} & \pythoninline{dict} & Additional information returned by the environment \\ \bottomrule
    \end{tabular}
\end{table}

The types of \pythoninline{observations} and \pythoninline{actions} depend on the corresponding spaces.

\end{document}